\documentclass{article}
\usepackage{hyperref}
\usepackage{spconf,amsmath,graphicx,hyperref}
\usepackage{booktabs}
\usepackage{multirow}
\usepackage{xcolor}

\ninept

\title{Relative Time Intervals Representation for Word-level Timestamping with Masked Training}
\name{Quanwei Tang$^1$, Zhiyu Tang$^2$, Xu Li$^3$, Dong Zhang$^{1,4\dagger}$
{\hypersetup{pdfborder={0 0 0}}
\thanks{
$^{\dagger}$Corresponding author.~
\href{dzhang@suda.edu.cn}{dzhang@suda.edu.cn}
}}
, Shoushan Li$^1$, Guodong Zhou$^1$}
\address{$^1$Soochow University $^2$ University of Queenland $^3$AISpeech Ltd $^4$Jiangsu Key Lab of Language Computing
}
\begin{document}
%
\maketitle
\begin{abstract}
Although Speech Large Language Models (SpeechLLMs) excel at speech understanding and generation, their capacity for fine-grained, temporally aligned outputs remains underexplored. Our work addresses this gap by enabling SpeechLLMs to jointly model speech content and temporal structure, effectively transforming them from ``content understanding machines" into ``temporal-aware content understanding machines". Specifically, we replace traditional absolute timestamps with relative timestamps, achieving a more compact vocabulary and stronger generalization capabilities. To efficiently infuse timestamp prediction ability into pre-trained large language models, we introduce a hybrid fine-tuning strategy: full-parameter fine-tuning of the timestamp-augmented embedding layer and language model head, combined with LoRA fine-tuning of the decoder layers. Moreover, we design a masked timestamp training objective, preventing the model from over-relying on ground-truth timestamps, and thereby enhancing robustness against noisy real-world annotations. Extensive experiments demonstrate that our approach achieves significant improvements in timestamp prediction accuracy while maintaining strong speech transcription performance. 
\footnote{https://github.com/tangquanwei/Timestamp-Aware-Speech-LLM}

\end{abstract}

\section{Introduction}

\begin{figure}[t]
    \centering
    \includegraphics[width=1\linewidth]{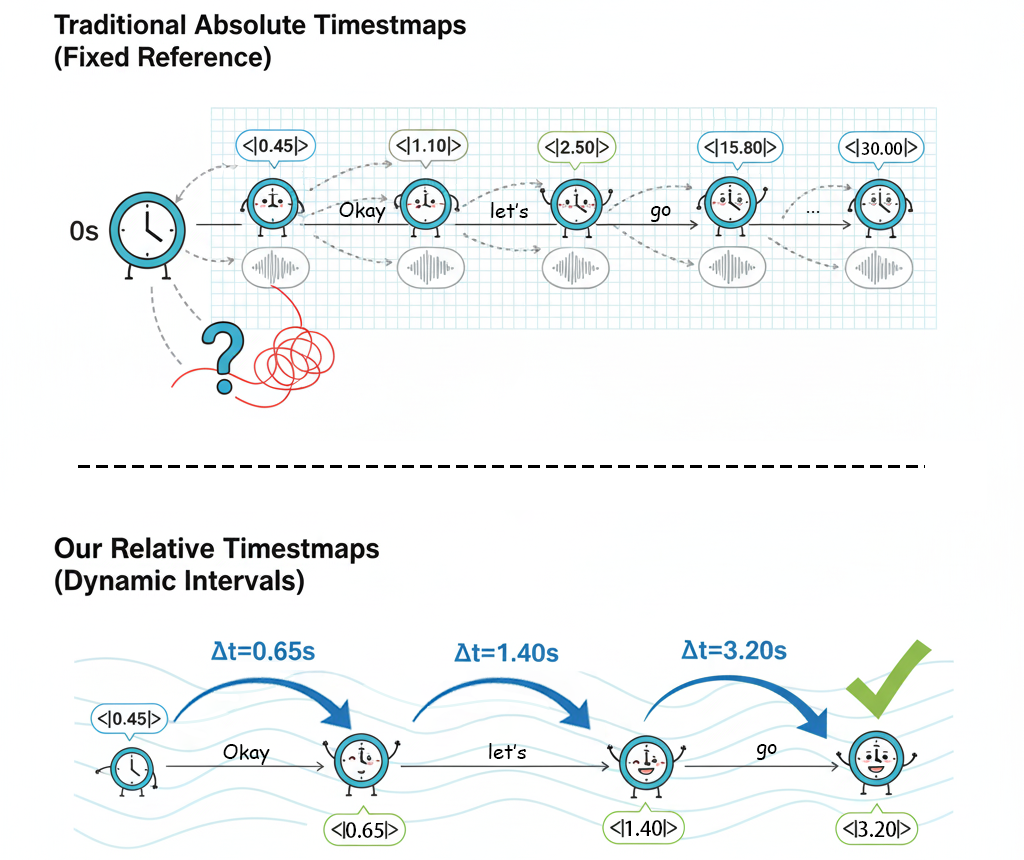}
    \caption{Traditional methods rely on fixed-reference absolute timestamps, often struggling with cumulative errors or precise synchronization. Our approach utilizes dynamic interval-based relative timestamps, which inherently model temporal relationships and offer improved robustness and accuracy.}
    \label{fig:intro}
\end{figure}


Recent research has explored the integration of large language models (LLMs) with speech processing \cite{rousso2024tradition, peng2024survey}. However, generating fine-grained, word-level timestamped transcription has not been fully explored yet \cite{papi2024leveraging,sivasankaran2025target}. This task demands not only accurate speech understanding but also precise temporal alignment between linguistic units and their acoustic realizations \cite{makishima2023joint,hu2025word}. A core challenge lies in how to represent and predict temporal structure efficiently within the constrained vocabulary and sequential decoding paradigm of LLMs \cite{yamasaki2023transcribing,yu2025unsupervised,tang-etal-2025-comprehensive, wu2025emotion}. Existing methods like Qwen2 Audio \cite{chu2024qwen2audiotechnicalreport} often employ absolute timestamps for modeling. e.g., \verb!<|0.45|>Okay<|1.10|>let's<|2.50|>go<|5.70| >!\small{...}
indicates that the word "Okay" starts at 0.45 seconds and ends at 1.10 seconds. While this representation is intuitive and facilitates evaluation, its dependence on an absolute temporal reference limits scalability and generalization over long-duration audio segments.

As shown in Figure \ref{fig:intro}, predicting absolute timestamps is prone to cumulative error propagation. Absolute timestamps are monotonically increasing, causing LLMs to learn a cumulative pattern: generating larger values by adding estimated segment durations to prior timestamps. Furthermore, this approach exhibits poor generalization to unseen temporal ranges. For instance, it fails to predict timestamps beyond the training duration (e.g., at 60 seconds when trained only on utterances up to 30 seconds). A major practical limitation lies in the prohibitively large vocabulary required for high temporal precision: achieving 0.01-second resolution over a 300-second audio clip requires 30,000 distinct timestamp tokens. This explosion in vocabulary significantly increases training complexity and computational overhead, making the approach unscalable for long-form speech processing.

\begin{figure*}[t]
    \centering
    \includegraphics[width=1\linewidth]{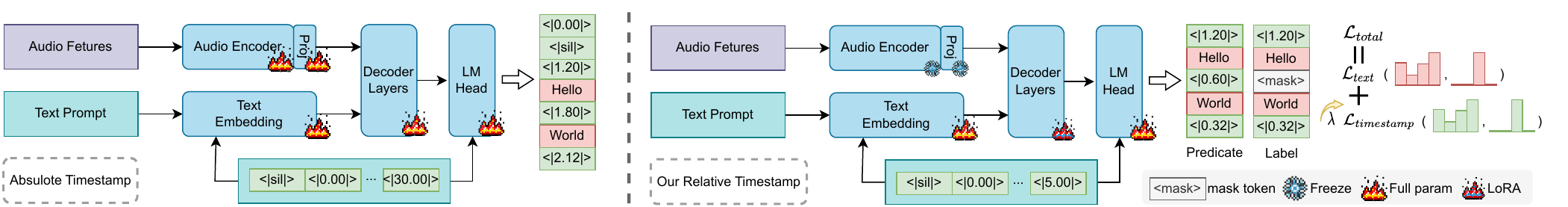}
    \caption{Architectural Comparison of an Absolute Timestamp-based Model (Left) and Our Proposed Relative Timestamp-based Model (Right). The traditional method models timestamps as independent absolute values and struggles with representing timestamps. In contrast, our approach utilizes a compact vocabulary of relative timestamps. By leveraging LoRA fine-tuning on the decoder, our model efficiently learns to jointly generate both text and temporal markers. To ensure accuracy and robustness, our training incorporates a combined loss function and timestamp regularization (token masking), which forces the model to learn a more resilient representation of temporal information.}
    \label{fig:method}
\end{figure*}
To address these limitations, we propose a relative timestamp representation, defined as the time interval between consecutive words. (e.g., \verb!<|0.65|>! for words aligned at \verb!<|0.45|>! and \verb!<|1.10|>!).
This method requires solely estimating the current segment's duration, thereby minimizing error susceptibility.
Only a limited time interval (such as 0.01 seconds to 5 seconds) needs to be learned, and by accumulating these intervals, any length of audio can be represented. Vocabulary only needs to contain limited time interval tokens, compared to absolute timestamps. Human speech perception naturally focuses on relative timing: we intuitively process how long after the previous word the next one occurs, rather than its absolute position in time. This relative interval modeling aligns naturally with the autoregressive generation paradigm of language models. Absolute timestamps force the model to learn an entirely new absolute value that is unrelated to the previously generated content, which is contrary to its natural generation mode. 

We make the following three key contributions:

1. \textbf{A Novel Relative Time Interval Representation for Word-Level Timestamping:} We introduce a relative timestamp representation that models inter-word intervals instead of absolute positions. This approach mitigates vocabulary explosion and temporal drift, enabling accurate timestamp prediction for arbitrarily long audio without length-dependent degradation.

2. \textbf{Advanced Regularization and Alignment Strategies:} We propose a regularization technique, \textit{Timestamp Masking}, which mitigates overfitting to perfect alignments, along with a dynamically weighted \textit{Joint Temporal Alignment  Loss}  that adaptively balances transcription accuracy and timestamp fidelity during training.

3. \textbf{Hybrid Parameter-Efficient Fine-Tuning:} We fully fine-tune only timestamp-specific modules (embedding layer and LM head), while adapting pre-trained decoder layers via LoRA. This role-aware strategy preserves linguistic knowledge with minimal parameter updates, aligning adaptation granularity with each module’s functional role.

\section{Methodology}

\subsection{Relative Timestamp Representation}

Unlike traditional methods that rely on an extensive vocabulary of absolute timestamps, our approach employs \textbf{relative timestamps} to represent temporal information. Each timestamp token signifies the time interval from the end of the preceding word to the current word. The core advantage of this method is its superior generalization and compact vocabulary. By learning a constrained set of time intervals, the model can cumulatively represent temporal information for audio of any length, eliminating the need to define a unique token for every possible absolute time point.

\subsection{LLMs-based Architecture and Fine-Tuning}

Our model architecture is built upon a pre-trained large speech model, which consists of an encoder for extracting speech features and a decoder for sequence generation.

As shown in Figure \ref{fig:method} (right), to effectively adapt our model for the new task of temporal alignment, we employ a hybrid fine-tuning strategy that combines full parameter updates for new components with parameter-efficient fine-tuning for backbone modules.

We apply full-parameter fine-tuning to the newly introduced timestamp embedding and the Language Model (LM) head. These modules are directly responsible for mapping the new timestamp tokens to a continuous vector space and generating them in the output sequence, respectively. Since their initial weights are random, a comprehensive update of all parameters is necessary for them to effectively learn and represent the new temporal-augmented vocabulary from scratch.

For the architecture of the backbone model, specifically the decoder layers, we utilize LoRA (Low-Rank Adaptation) \cite{hu2022lora}. This strategic choice is driven by both computational efficiency and the need for targeted adaptation. Instead of updating all parameters, LoRA freezes the original weights and injects small, trainable low-rank matrices. This approach drastically reduces the number of trainable parameters, leading to significant savings in computational resources. Furthermore, by fine-tuning the decoder with LoRA, the model can efficiently learn to insert corresponding temporal markers when generating specific text content, without interfering with the powerful, pre-trained understanding of speech features. This hybrid approach ensures computational efficiency while achieving precise control over the model's generative behavior for robust and accurate temporal alignment.

\subsection{Training Strategies for Temporal Alignment}

To ensure that our model generates accurate and robust timestamps, we designed two key training strategies:

\subsubsection{Joint Temporal Alignment Loss}

Our training objective is to minimize a joint loss function composed of two components, which balances the weights between text transcription and timestamp generation:
$$L_{total} = L_{text} + \lambda L_{timestamp}$$
where $L_{text}$ is a standard cross-entropy loss measuring the discrepancy between the model's generated text tokens and the ground-truth tokens. Similarly, $L_{timestamp}$ is a cross-entropy loss that evaluates the prediction accuracy of the discrete timestamp tokens.

The hyperparameter $\lambda$ serves as a weighting factor between the two loss components. We employ a dynamic weighting strategy: in the early training stages, we set a smaller weight ($\lambda=1$) to allow the model to prioritize learning the speech-to-text mapping. As training progresses, we gradually increase the value of $\lambda$ to encourage more precise temporal alignment. Specifically, for each training epoch, we increment the value of $\lambda$ by 1.

\begin{table*}[t]
    \centering
    \footnotesize
    \caption{Performance comparison of different models on timestamp prediction. Precision and Recall are in percentages, and the average timestamp difference is in milliseconds. '-' denotes the model does not support Chinese speech.}
    \label{tab:timestamp_comparison}
    \begin{tabular}{@{}ll rrr rrr@{}}
        \toprule
        \multirow{2}{*}{\textbf{Model}} & \multirow{2}{*}{\textbf{Tolerance (ms)}} & \multicolumn{3}{c}{\textbf{Librispeech}} & \multicolumn{3}{c}{\textbf{Wenet-Meeting}} \\
        \cmidrule(lr){3-5} \cmidrule(lr){6-8}
        & & \textit{P (\%)} & \textit{R (\%)} & \textit{Avg. Diff (ms)} & \textit{P(\%)} & \textit{R (\%)} & \textit{Avg. Diff (ms)} \\
        \midrule
        \multirow{3}{*}{Qwen2-Audio \cite{chu2024qwen2audiotechnicalreport}} & 80 & 0.00 & 0.00 & 1031.23 & 60.81 & 56.06 & 59.03 \\
                    & 160 & 0.01 & 0.00 & 1031.18 & 76.07 & 70.13 & 49.11 \\
                    & 240 & 0.16 & 0.05 & 1030.42 & 82.30 & 75.87 & 43.07 \\
        \midrule
        \multirow{3}{*}{WhisperTimestamed \cite{whisper}}     & 80 & 2.70 & 2.36 & 131.92 & 3.54 & 1.51 & 164.09 \\
                    & 160 & 10.09 & 8.83 & 120.52 & 24.30 & 10.38 & 134.03 \\
                    & 240 & 19.90 & 17.42 & 101.14 & 43.97 & 18.79 & 102.48 \\
        \midrule
       \multirow{3}{*}{SenseVoiceSmall \cite{SenseVoice-Small}}  & 80 & 1.10 & 1.10 & 176.38 & 28.52 & 27.96 & 80.80 \\
                    & 160 & 5.20 & 5.19 & 166.71 & 71.40 & 69.99 & 45.26 \\
                    & 240 & 12.88 & 12.86 & 148.59 & 82.82 & 81.19 & 32.47 \\
        \midrule
       \multirow{3}{*}{Canary \cite{hu25e_interspeech}}  & 80 & 35.51 & 35.27  & 444.30  & - & - & - \\
                                & 160 & 71.23 & 70.74 & 268.18 & - & - & - \\
                                & 240 & \textbf{84.57} & \textbf{83.99} & 200.47 & - & - & - \\
        \midrule
        \multirow{3}{*}{Absolute Timestamp}      & 80 & 12.30 & 5.51 & 159.05 & 45.14 & 42.52 & 186.73 \\
                    & 160 & 17.03 & 7.63 & 156.58 & 62.99 & 59.34 & 168.29 \\
                    & 240 & 21.44 & 9.61 & 152.51 & 72.41 & 68.22 & 152.17 \\
        \midrule
        \multirow{3}{*}{\textbf{Relative Timestamp (Ours)}}      & 80 & \textbf{40.49} & \textbf{38.44} & 145.72 & \textbf{61.07} & \textbf{60.62} & 55.80 \\
                    & 160 & \textbf{77.74} & \textbf{75.24} & 139.21 & \textbf{78.04} & \textbf{75.46} & 42.45 \\
                    & 240 & 83.65 & 82.64 & 127.91 & \textbf{91.13} & \textbf{86.88} & \textbf{30.34} \\
        \bottomrule
    \end{tabular}
    \vspace{0.5em} 
\end{table*}

\subsubsection{Timestamp Masking for Regularization}

To mitigate overfitting to ground-truth during training, we propose a timestamp masking technique. The core idea involves randomly replacing a subset of timestamp tokens with a [MASK] token. During training, the previous timestamp token is masked, the model cannot reference its ground-truth value, and must instead infer the current timestamp from speech content and historical tokens, thereby enhancing robustness to inaccurate annotations.
Similar to BERT's masked language modeling, this random masking encourages the model to leverage varying input components across iterations, preventing over-reliance on specific timestamps.

\section{Experiments}

\subsection{Experimental Setup}

\textbf{Datasets.} We conduct experiments on five benchmark datasets:
AISHELL-1: A Mandarin corpus with $\sim$178 hours of speech from 400 speakers \cite{AISHELL-1}.
AISHELL-2: A larger Mandarin dataset containing $\sim$1,000 hours from 1,991 speakers in diverse acoustic environments \cite{du2018aishell2transformingmandarinasr}.
Wenet Meeting: A multi-domain Mandarin corpus consisting of 10000+ hours of high-quality labeled speech \cite{zhang2022wenetspeech}.
LibriSpeech: An English read-speech corpus with $\sim$1,000 hours of audiobooks, widely adopted for ASR evaluation \cite{Librispeech}.
Common Voice (English subset): A crowd-sourced corpus with diverse accents and recording conditions \cite{ardila-etal-2020-common}.

\textbf{Baseline Models.} We compare our proposed method against several state-of-the-art models.
Qwen2-Audio: A large language model with integrated audio understanding capabilities, trained using a large amount of diverse data and multiple training methodologies \cite{chu2024qwen2audiotechnicalreport}.
WhisperTimestamped: This model utilizes Whisper-large-v3, a robust, multilingual ASR model trained on approximately 680,000 hours of weakly supervised audio data \cite{whisper}.
SenseVoice-Small: A lightweight and efficient ASR system, which is trained on over 400,000 hours of data and supports more than 50 languages \cite{SenseVoice-Small}.
Canary: Employs a data-driven approach to enable word-level timestamp prediction and supports multiple languages other than Chinese \cite{hu25e_interspeech}.

\textbf{Metric.} 
For timestamp prediction, we introduce the concept of time tolerance. Due to the inherent ambiguity of speech boundaries, a predicted timestamp is considered correct if its difference from the ground-truth timestamp falls within a predefined tolerance \cite{rastorgueva2023nemo,kim2024automatic,ji2025pathological}.
We use Precision and Recall to evaluate the accuracy of these timestamps:
Precision measures the proportion of correctly predicted timestamps among all predicted timestamps.
Recall measures the proportion of correctly predicted timestamps among all ground-truth timestamps.
Additionally, we use the Average Time Difference to quantify the temporal accuracy, defined as:
$$
Avg. Diff= \frac{\sum_{i=1}^N |timestamp_{gt}^i-timestamp_{pred}^i|}{N}
$$

For ASR analysis, we use Word Error Rate (WER) as the primary metric \cite{xu2023efficient}.

\begin{table*}[ht]
    \centering
    \footnotesize
    \caption{Ablation study on timestamp prediction performance at a tolerance of 240 ms (WER \% and Precision/Recall) across different model configurations.}
    \label{tab:ablation_study}
    \begin{tabular}{@{}l ccc ccc@{}}
        \toprule
        \multirow{2}{*}{\textbf{Configuration}} & \multicolumn{3}{c}{\textbf{AISHELL-2 iOS (Chinese)}} & \multicolumn{3}{c}{\textbf{Common Voice (English)}} \\
        \cmidrule(lr){2-4} \cmidrule(lr){5-7}
        & \textbf{WER (\%)} & \textbf{Precision} & \textbf{Recall} & \textbf{WER (\%)} & \textbf{Precision} & \textbf{Recall} \\
        \midrule
        \textbf{Absolute Timestamp} & 2.87 & 0.9544 & 0.9546 & 16.41 & 0.7861 & 0.7838 \\
        \quad - TS Loss             & 2.96 & 0.9506 & 0.9447 & 18.63 & 0.7428 & 0.7407 \\
        \midrule
        \textbf{Relative Timestamp} & \textbf{2.15} & \textbf{0.9763} & \textbf{0.9634} & \textbf{11.63} & \textbf{0.8770} & \textbf{0.8546} \\
        \quad - TS Loss             & 2.31 & 0.9715 & 0.9548 & 12.66 & 0.8712 & 0.8463 \\
        \quad - Timestamp Masking      & 2.56 & 0.9658 & 0.9578 & 14.47 & 0.8067 & 0.7853 \\
        \bottomrule
    \end{tabular}
\end{table*}

\textbf{Implementation Details.} 
Our training methodology extends the FireRedASR-LLM's \cite{xu2025fireredasr} architecture to support timestamp outputs. 
This model is built upon a conformer audio encoder and the Qwen2-7B-Instruct backbone LLM. Crucially, the native model cannot generate timestamps.

The model is trained on a cluster of 24 x Ascend 910B (64G) NPUs for 7k steps, with each NPU handling a batch duration of 500 seconds. Setting $\lambda$ as 1 increases by 1 for each epoch, and empirically setting masking probability as 10\% from the second epoch.
For training, we use the AdamW optimizer with a low learning rate of $5 \times 10 ^{-6}$ and a WarmupCosineLR scheduler to ensure stable convergence. The entire process utilizes bf16 mixed-precision training for enhanced efficiency.
To enable timestamp generation, we incorporate new tokens into the LLM's vocabulary:
Absolute Timestamp Pattern: We add tokens from \verb!<|0.00|>! to \verb!<|30.00|>!, representing a time range of 0 to 30 seconds.
Relative Timestamp Pattern: We add tokens from \verb!<|0.00|>! to \verb!<|5.00|>!, representing time differences up to 5 seconds.
Unlike traditional methods that require a special instruction token \verb!<|timestamp|>! Our model is naturally prompted for this task by a simple, intuitive command like "Speech to text with timestamp."
Our training regimen was confined to a pair of widely-used datasets: the AISHELL-2 corpus \cite{du2018aishell2transformingmandarinasr} for Mandarin and the English subset of the Common Voice dataset \cite{ardila-etal-2020-common}.

\subsection{Main Results}

Table \ref{tab:timestamp_comparison} compares the timestamp prediction performance of various models across the LibriSpeech and Wenet-Meeting datasets.

Our Relative Timestamp method achieved the highest Precision and Recall scores across all tolerance levels on both datasets. For instance, at a 240 ms tolerance, our model reached an impressive 91.13\% Precision and 86.88\% Recall on Wenet-Meeting, substantially outperforming all baselines.
Crucially, our model also exhibited the best temporal accuracy. With an average timestamp difference of just 30.34 ms at a 240 ms tolerance on Wenet-Meeting, our predictions were the closest to the ground truth.

In contrast, Canary performs worse than our method at low tolerance levels but slightly better at high tolerance levels. This discrepancy can be attributed to the fact that Canary was trained on the Librispeech dataset, whereas our model was not.Qwen2-Audio failed on LibriSpeech, with Precision and Recall scores near zero, indicating its inability to handle timestamps on this English corpus.WhisperTimestamped and SenseVoiceSmall generally delivered lower performance, especially at lower tolerances, with their results falling far behind ours.
We also conducted an ASR experiment, and the WER results \ref{tab:wer_compact} showed that our model outperforms all baselines on all datasets.

In conclusion, this table definitively proves that our Relative Timestamp method sets a new benchmark for timestamp prediction. By focusing on relative temporal relationships, our approach achieves superior accuracy and precision, validating its potential for building high-performance speech language models.

\subsection{Ablation study}

Table \ref{tab:ablation_study} presents a detailed ablation study on the performance of different model configurations. Our Relative Timestamp model consistently achieves the best performance across all metrics. The consistently higher Precision and Recall scores further confirm our model’s superior ability to not only transcribe words accurately but also to predict their timestamps with greater precision.

To dissect our method’s key contributions, we conducted further ablations:

\textbf{Impact of Timestamp Loss (TS Loss):} Removing the timestamp loss term (- TS Loss) results in a performance drop for both baseline and our models, as evidenced by an increase in WER and a decrease in Precision/Recall. This highlights the critical role of TS Loss in guiding the model to learn accurate timestamp prediction, preventing it from solely focusing on the text sequence generation task.

\textbf{Impact of Timestamp Mask:} The ablation study on our model reveals a significant performance degradation when the timestamp mask is removed (- Timestamp Mask). The WER increases from 2.15\% to 2.56\% on AISHELL-2 iOS and from 11.63\% to 14.47\% on Common Voice. This compellingly demonstrates that the Timestamp Mask is crucial for the training process. It helps the model focus on valid timestamp predictions and filters out noise, ensuring the robustness of our relative timestamp.

\subsection{Analysis of ASR}
\label{asr_wer}
As depicted in Table \ref{tab:wer_compact}, a comprehensive performance comparison based on Word Error Rate (WER) is presented across five diverse datasets. Our proposed method, Relative Timestamp, consistently demonstrates superior performance across all evaluated datasets. This is visually confirmed by its polygon having the largest area, signifying its robust generalization and leading performance across multilingual and domain-specific tasks. Specifically, our model achieves WERs of 1.26\% and 2.15\% on the Chinese datasets AISHELL-1 (AS-1) and AISHELL-2 (AS-2), respectively, significantly outperforming all baselines. On the more challenging Wenet Meeting (Wenet) dataset, our model's WER of 5.56\% establishes a substantial lead. Furthermore, our approach also secures the best results on the English datasets LibriSpeech (Libri) (2.78\%) and Common Voice (CV) (11.63\%).

\begin{table}[h]
    \centering
    \caption{WER ($\%$) comparison. Lower is better ($\downarrow$).}
    \label{tab:wer_compact}
    \resizebox{\columnwidth}{!}{%
        \begin{tabular}{l ccccc}
            \toprule
            \textbf{Method} & \textbf{AS-1} & \textbf{AS-2} & \textbf{Wenet} & \textbf{Libri} & \textbf{CV} \\
             & \tiny{(CN)} & \tiny{(CN)} & \tiny{(CN)} & \tiny{(EN)} & \tiny{(EN)} \\
            \midrule
            Qwen2 Audio & 1.62 & 3.38 & 21.99 & 43.32 & 74.37 \\
            SenseVoiceSmall & 3.10 & 3.87 & 7.54 & 3.80 & 15.97 \\
            WhisperTimestamped & 12.46 & 8.08 & 56.66 & 20.50 & 38.09 \\
            Absolute Timestamp & 1.38 & 2.87 & 11.30 & 9.11 & 16.41 \\
            \midrule
            \textbf{Ours (Relative Timestamp)} & \textbf{1.26} & \textbf{2.15} & \textbf{5.56} & \textbf{2.78} & \textbf{11.63} \\
            \bottomrule
        \end{tabular}%
    }
\end{table}




\section{Conclusion}
We propose a novel framework aimed at addressing the core challenges of existing speech models in precise temporal sequence understanding. We have gone beyond the simple method of adding timestamp tags to LLMs, instead focusing on endowing the model with fine-grained, aligned multi-modal spatiotemporal understanding capabilities, thereby upgrading LLMs from ``content understanding machines" to ``temporal-aware content understanding machines".

\section{Acknowledgements}

This was supported by Jiangsu Province Frontier Program Project (BF2025036), Hong Kong RGC grant GRF \#15611021, NSFC grant (No. 62376178), and Jiangsu Key Lab of Language Computing.

\vfill\pagebreak
\bibliographystyle{IEEEbib}
\bibliography{refs}

@inproceedings{xu2023efficient,
  title={Efficient sequence transduction by jointly predicting tokens and durations},
  author={Xu, Hainan and Jia, Fei and Majumdar, Somshubra and Huang, He and Watanabe, Shinji and Ginsburg, Boris},
  booktitle={International Conference on Machine Learning},
  pages={38462--38484},
  year={2023},
  organization={PMLR}
}

@inproceedings{rastorgueva2023nemo,
  title={Nemo forced aligner and its application to word alignment for subtitle generation},
  author={Rastorgueva, Elena and Lavrukhin, Vitaly and Ginsburg, Boris},
  booktitle={Proc. Interspeech},
  year={2023}
}

@article{kim2024automatic,
  title={Automatic recognition of second language speech-in-noise},
  author={Kim, Seung-Eun and Chernyak, Bronya R and Seleznova, Olga and Keshet, Joseph and Goldrick, Matthew and Bradlow, Ann R},
  journal={JASA Express Letters},
  volume={4},
  number={2},
  year={2024},
  publisher={AIP Publishing}
}

@inproceedings{yu2025unsupervised,
  title={Unsupervised Speech-text word-level alignment with Dynamic Programming},
  author={Yu, Tianshu and Gong, Zihan and Tan, Minghuan and Chen, Guhong and Yang, Min},
  booktitle={Findings of the Association for Computational Linguistics: NAACL 2025},
  pages={2323--2334},
  year={2025}
}

@inproceedings{yamasaki2023transcribing,
  title={Transcribing and aligning conversational speech: A hybrid pipeline applied to french conversations},
  author={Yamasaki, Hiroyoshi and Louradour, J{\'e}r{\^o}me and Hunter, Julie and Pr{\'e}vot, Laurent},
  booktitle={2023 IEEE Automatic Speech Recognition and Understanding Workshop (ASRU)},
  pages={1--6},
  year={2023},
  organization={IEEE}
}

@inproceedings{sivasankaran2025target,
  title={Target word activity detector: An approach to obtain ASR word boundaries without lexicon},
  author={Sivasankaran, Sunit and Sun, Eric and Li, Jinyu and Huang, Yan and Pan, Jing},
  booktitle={ICASSP 2025},
  pages={1--5},
  year={2025},
  organization={IEEE}
}

@inproceedings{papi2024leveraging,
  title={Leveraging timestamp information for serialized joint streaming recognition and translation},
  author={Papi, Sara and Wang, Peidong and Chen, Junkun and Xue, Jian and Kanda, Naoyuki and Li, Jinyu and Gaur, Yashesh},
  booktitle={ICASSP 2024-2024},
  pages={10381--10385},
  year={2024},
  organization={IEEE}
}

@article{rousso2024tradition,
  title={Tradition or innovation: A comparison of modern asr methods for forced alignment},
  author={Rousso, Rotem and Cohen, Eyal and Keshet, Joseph and Chodroff, Eleanor},
  journal={arXiv preprint arXiv:2406.19363},
  year={2024}
}

@article{hu2025word,
  title={Word Level Timestamp Generation for Automatic Speech Recognition and Translation},
  author={Hu, Ke and Puvvada, Krishna and Rastorgueva, Elena and Chen, Zhehuai and Huang, He and Ding, Shuoyang and Dhawan, Kunal and Xu, Hainan and Balam, Jagadeesh and Ginsburg, Boris},
  journal={arXiv preprint arXiv:2505.15646},
  year={2025}
}

@inproceedings{makishima2023joint,
  title={Joint autoregressive modeling of end-to-end multi-talker overlapped speech recognition and utterance-level timestamp prediction},
  author={Makishima, Naoki and Suzuki, Keita and Suzuki, Satoshi and Ando, Atsushi and Masumura, Ryo},
  booktitle={Proc. INTERSPEECH, 2023},
  pages={2913--2917},
  year={2023}
}

@inproceedings{AISHELL-1,
  author = {Bu, Hui and Du, Jiayu and Na, Xingyu and Wu, Bengu and Zheng, Hao},
  booktitle = {O-COCOSDA},
  ee = {https://doi.org/10.1109/ICSDA.2017.8384449},
  isbn = {978-1-5386-3333-5},
  pages = {1-5},
  publisher = {IEEE},
  title = {AISHELL-1: An open-source Mandarin speech corpus and a speech recognition baseline.},
  url = {http://dblp.uni-trier.de/db/conf/ococosda/ococosda2017.html#BuDNWZ17},
  year = 2017
}

@misc{du2018aishell2transformingmandarinasr,
      title={AISHELL-2: Transforming Mandarin ASR Research Into Industrial Scale}, 
      author={Jiayu Du and Xingyu Na and Xuechen Liu and Hui Bu},
      year={2018},
      eprint={1808.10583},
      archivePrefix={arXiv},
      primaryClass={cs.CL},
      url={https://arxiv.org/abs/1808.10583}, 
}

@inproceedings{Librispeech,
  author = {Panayotov, Vassil and Chen, Guoguo and Povey, Daniel and Khudanpur, Sanjeev},
  booktitle = {ICASSP 2015 },
  description = {Librispeech: An ASR corpus based on public domain audio books | IEEE Conference Publication | IEEE Xplore},
  doi = {10.1109/ICASSP.2015.7178964},
  issn = {2379-190X},
  month = {April},
  pages = {5206-5210},
  title = {Librispeech: An ASR corpus based on public domain audio books},
  url = {https://ieeexplore.ieee.org/document/7178964},
  year = 2015
}

@inproceedings{tang-etal-2025-comprehensive,
    title = "A Comprehensive Graph Framework for Question Answering with Mode-Seeking Preference Alignment",
    author = "Tang, Quanwei  and
      Lee, Sophia Yat Mei  and
      Wu, Junshuang  and
      Zhang, Dong  and
      Li, Shoushan  and
      Cambria, Erik  and
      Zhou, Guodong",
    editor = "Che, Wanxiang  and
      Nabende, Joyce  and
      Shutova, Ekaterina  and
      Pilehvar, Mohammad Taher",
    booktitle = "Findings of ACL 2025",
    month = jul,
    year = "2025",
    address = "Vienna, Austria",
    publisher = "Association for Computational Linguistics",
    url = "https://aclanthology.org/2025.findings-acl.1108/",
    doi = "10.18653/v1/2025.findings-acl.1108",
    pages = "21504--21523",
    ISBN = "979-8-89176-256-5"
}

@inproceedings{ji2025pathological,
  title={Pathological Section Staining Transferring with Tailored Metric-based Model Selection},
  author={Ji, Yiming and Zhu, Suyang and Zhang, Dong and Li, Shoushan},
  booktitle={ICASSP 2025},
  pages={1--5},
  year={2025},
  organization={IEEE}
}

@inproceedings{wu2025emotion,
  title={Emotion across modalities and cultures: Multilingual multimodal emotion-cause analysis with memory-inspired framework},
  author={Wu, Dan and Ju, Xincheng and Zhang, Dong and Li, Shoushan and Cambria, Erik and Zhou, Guodong},
  booktitle={Proceedings of the 33rd ACM International Conference on Multimedia},
  pages={5775--5783},
  year={2025}
}

@inproceedings{ardila-etal-2020-common,
    title = "Common Voice: A Massively-Multilingual Speech Corpus",
    author = "Ardila, Rosana  and
      Branson, Megan  and
      Davis, Kelly  and
      Kohler, Michael  and
      Meyer, Josh  and
      Henretty, Michael  and
      Morais, Reuben  and
      Saunders, Lindsay  and
      Tyers, Francis  and
      Weber, Gregor",
    booktitle = "Proceedings of the Twelfth Language Resources and Evaluation Conference",
    month = "May",
    year = "2020",
    address = "Marseille, France",
    publisher = "European Language Resources Association",
    url = "https://aclanthology.org/2020.lrec-1.520/",
    pages = "4218--4222",
    language = "eng",
    ISBN = "979-10-95546-34-4"
}

@misc{whisper,
      title={Robust Speech Recognition via Large-Scale Weak Supervision}, 
      author={Alec Radford and Jong Wook Kim and Tao Xu and Greg Brockman and Christine McLeavey and Ilya Sutskever},
      year={2022},
      eprint={2212.04356},
      archivePrefix={arXiv},
      primaryClass={eess.AS},
      url={https://arxiv.org/abs/2212.04356}, 
}

@article{SenseVoice-Small,
  author       = {Keyu An and
                  Qian Chen and
                  Chong Deng and
                  Zhihao Du and
                  Changfeng Gao and
                  Zhifu Gao and
                  Yue Gu and
                  Ting He and
                  Hangrui Hu and
                  Kai Hu and
                  Shengpeng Ji and
                  Yabin Li and
                  Zerui Li and
                  Heng Lu and
                  Haoneng Luo and
                  Xiang Lv and
                  Bin Ma and
                  Ziyang Ma and
                  Chongjia Ni and
                  Changhe Song and
                  Jiaqi Shi and
                  Xian Shi and
                  Hao Wang and
                  Wen Wang and
                  Yuxuan Wang and
                  Zhangyu Xiao and
                  Zhijie Yan and
                  Yexin Yang and
                  Bin Zhang and
                  Qinglin Zhang and
                  Shiliang Zhang and
                  Nan Zhao and
                  Siqi Zheng},
  title        = {FunAudioLLM: Voice Understanding and Generation Foundation Models
                  for Natural Interaction Between Humans and LLMs},
  journal      = {CoRR},
  volume       = {abs/2407.04051},
  year         = {2024},
  url          = {https://doi.org/10.48550/arXiv.2407.04051},
  doi          = {10.48550/ARXIV.2407.04051},
  eprinttype    = {arXiv},
  eprint       = {2407.04051},
  bibsource    = {dblp computer science bibliography, https://dblp.org}
}

@misc{chu2024qwen2audiotechnicalreport,
      title={Qwen2-Audio Technical Report}, 
      author={Yunfei Chu and Jin Xu and Qian Yang and Haojie Wei and Xipin Wei and Zhifang Guo and Yichong Leng and Yuanjun Lv and Jinzheng He and Junyang Lin and Chang Zhou and Jingren Zhou},
      year={2024},
      eprint={2407.10759},
      archivePrefix={arXiv},
      primaryClass={eess.AS},
      url={https://arxiv.org/abs/2407.10759}, 
}

@article{xu2025fireredasr,
  title={FireRedASR: Open-Source Industrial-Grade Mandarin Speech Recognition Models from Encoder-Decoder to LLM Integration},
  author={Xu, Kai-Tuo and Xie, Feng-Long and Tang, Xu and Hu, Yao},
  journal={arXiv preprint arXiv:2501.14350},
  year={2025}
}

@article{hu2022lora,
  title={Lora: Low-rank adaptation of large language models.},
  author={Hu, Edward J and Shen, Yelong and Wallis, Phillip and Allen-Zhu, Zeyuan and Li, Yuanzhi and Wang, Shean and Wang, Lu and Chen, Weizhu and others},
  journal={ICLR},
  volume={1},
  number={2},
  pages={3},
  year={2022}
}

@inproceedings{hu25e_interspeech,
  title     = {{Word Level Timestamp Generation for Automatic Speech Recognition and Translation}},
  author    = {Ke Hu and Krishna Puvvada and Elena Rastorgueva and Zhehuai Chen and He Huang and Shuoyang Ding and Kunal Dhawan and Hainan Xu and Jagadeesh Balam and Boris Ginsburg},
  year      = {2025},
  booktitle = {{Interspeech 2025}},
  pages     = {2565--2569},
  doi       = {10.21437/Interspeech.2025-869},
  issn      = {2958-1796},
}

@inproceedings{zhang2022wenetspeech,
  title={Wenetspeech: A 10000+ hours multi-domain mandarin corpus for speech recognition},
  author={Zhang, Binbin and Lv, Hang and Guo, Pengcheng and Shao, Qijie and Yang, Chao and Xie, Lei and Xu, Xin and Bu, Hui and Chen, Xiaoyu and Zeng, Chenchen and others},
  booktitle={ICASSP 2022},
  pages={6182--6186},
  year={2022},
  organization={IEEE}
}

@article{peng2024survey,
  title={A survey on speech large language models},
  author={Peng, Jing and Wang, Yucheng and Xi, Yu and Li, Xu and Zhang, Xizhuo and Yu, Kai},
  journal={arXiv e-prints},
  pages={arXiv--2410},
  year={2024}
}

\end{document}


\appendix

\section{Appendix}

\subsection{Case Study}

\begin{figure*}[h]
    \centering
    \includegraphics[width=1\linewidth]{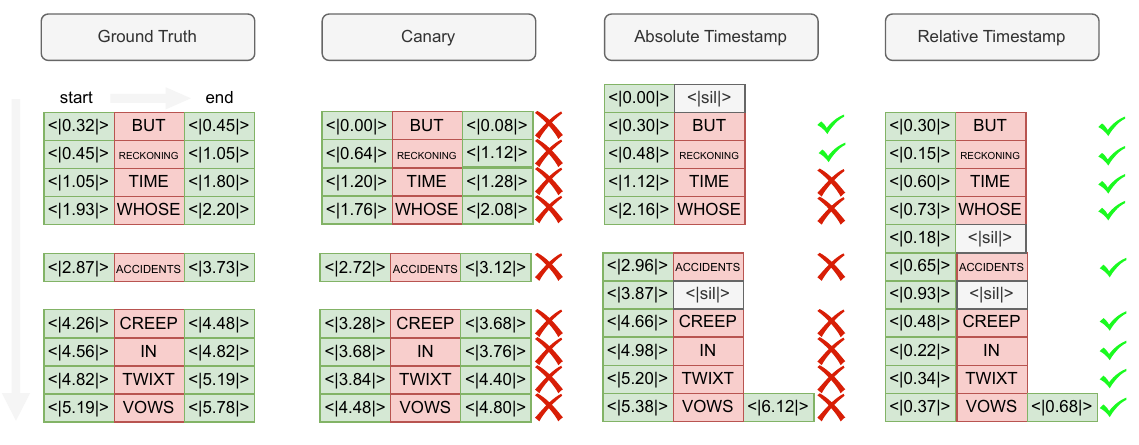}
    \caption{A Case from Librispeech, }
    \label{fig:case_study}
\end{figure*}

From the figure \ref{fig:case_study}, the second column shows Canary's outputs, where each word is in between the two timestamps token. However, in the Absolute Timestamp column, two adjacent words share one timestamp token. Our Relative Timestamp method, on the other hand, marks the duration of each word relative to the previous one. As Canary begins with an inaccurate timestamp, none of its subsequent timestamps are accurate. The Absolute Timestamp method predicts accurately at first but fails later on. Our Relative Timestamp method, however, keeps the error within an acceptable tolerance.

\subsection{Analysis of ASR}
\label{asr_wer}
As depicted in Fig. \ref{fig:wer_radar_chart}, a comprehensive performance comparison based on Word Error Rate (WER) is presented across five diverse datasets. Our proposed method, (Ours) Relative Timestamp, consistently demonstrates superior performance across all evaluated datasets. This is visually confirmed by its polygon having the smallest area, signifying its robust generalization and leading performance across multilingual and domain-specific tasks. Specifically, our model achieves WERs of 1.26\% and 2.15\% on the Chinese datasets AISHELL-1 and AISHELL-2, respectively, significantly outperforming all baselines. On the more challenging Wenet Meeting dataset, our model's WER of 5.56\% establishes a substantial lead. Furthermore, our approach also secures the best results on the English datasets LibriSpeech (2.78\%) and Common Voice (11.63\%).
 \begin{figure}[h]
     \centering
     \includegraphics[width=0.8\linewidth]{WER.png}
     \caption{Word Error Rate (WER \%) comparison of various models across five representative speech datasets. A smaller polygon area indicates lower WER and better performance.}
    \label{fig:wer_radar_chart}
 \end{figure}

\subsection{Implementation Details.}
The Base model FireRedASR-LLM \cite{xu2025fireredasr} is built upon a conformer audio encoder and the Qwen2-7B-Instruct backbone LLM. Crucially, the native model lacks the capability to generate timestamps.
Our training methodology extends this architecture to support timestamp output. We employ DeepSpeed with a ZeRO Stage 2 configuration to optimize memory usage and training speed. The model is trained on a cluster of 24 x Ascend 910B (64G) NPUs for 7k steps, with each NPU handling a batch duration of 500 seconds.Increasing $\lambda$ from 1 to 2 for two epochs, and setting masking probability as 10\% from epoch 2.
For training, we use the AdamW optimizer with a low learning rate of $5 \times 10 ^{-6}$ and a WarmupCosineLR scheduler to ensure stable convergence. The entire process utilizes bf16 mixed-precision training for enhanced efficiency.
To enable timestamp generation, we incorporate new tokens into the LLM's vocabulary:
Absolute Timestamp Pattern: We add tokens from \verb!<|0.00|>! to \verb!<|30.00|>!, representing a time range of 0 to 30 seconds.
Relative Timestamp Pattern: We add tokens from \verb!<|0.00|>! to \verb!<|5.00|>!, representing time differences up to 5 seconds.
Unlike traditional methods that require a special instruction token \verb!<|timestamp|>! Our model is naturally prompted for this task by a simple, intuitive command like "Speech to text with timestamp."
Our training regimen was confined to a pair of widely-used datasets: the AISHELL-2 corpus for Mandarin and the English subset of the Common Voice dataset .

